\documentclass[amsmath,amssymb,floatfix,
 aps, pre, twocolumn, superscriptaddress
]{revtex4-2}

\usepackage{amssymb}
\usepackage{amsmath}
\usepackage{placeins}
\usepackage{subfig}
\usepackage[colorlinks]{hyperref}
\usepackage{standalone}
\usepackage{graphics,graphicx,color, calc}
\usepackage[utf8]{inputenc}
\usepackage{amsfonts}
\usepackage[ruled,vlined]{algorithm2e}
\usepackage{booktabs}
\usepackage{soul}
\usepackage{url}
\usepackage[english]{babel}
\usepackage{mathrsfs}
\usepackage[mathcal]{eucal}
\usepackage[normalem]{ulem}
\usepackage{float}

\begin{document}

\title{SolarTformer: A Transformer Based Deep Learning Approach for Short Term Solar Power Forecasting}

\author{Ankan Basu}
\email[Corresponding Author: ]{ankanb.cse.pg@jadavpuruniversity.in}
\affiliation{Positional Astronomy Centre, India Meteorological Department, Kolkata, India}
\affiliation{Department of Computer Science and Engineering, Jadavpur University, Kolkata, India}

\author{Jyotiraditya Roy}
\thanks{Equal Contribution}
\affiliation{Department of Computer Science and Engineering (Artificial Intelligence and Machine Learning), Heritage Institute of Technology, Kolkata, India}

\author{Aditya Datta}
\thanks{Equal Contribution}
\affiliation{Department of Computer Science and Engineering (Artificial Intelligence and Machine Learning), Heritage Institute of Technology, Kolkata, India}

\author{Prayas Sanyal}
\thanks{Equal Contribution}
\affiliation{Department of Electronics and Communication Engineering, Heritage Institute of Technology, Kolkata, India}

\author{Sumanta Banerjee}
\affiliation{Department of Mechanical Engineering, Heritage Institute of Technology, Kolkata, India}

\begin{abstract}
 Accurate forecasting of solar power output is essential for efficient integration of renewable energy into the grid. In this study, an attention-based deep learning model, inspired by transformer architecture, is used for short-term solar power forecasting. Our proposed model, "SolarTformer", is designed to predict solar power output from meteorological data. Unlike traditional models, SolarTformer leverages self-attention mechanisms to effectively capture temporal dependencies and spatial variability in solar irradiance. In addition, the proposed methodology includes feeding power station-specific metadata into the model, which helps to generalize between power stations located at different locations and with different panel configurations and in different seasons. Our experiments demonstrate that SolarTformer significantly outperforms previous models on the same data set. In particular, the model exhibits strong performance on both clear and cloudy days, indicating high robustness and generalizability. These findings highlight the potential of attention-based architectures in enhancing the accuracy of solar forecasting, contributing to a more reliable management of renewable energy.
\end{abstract}

\maketitle

\noindent \textbf{Keywords:} Solar energy; Transformers; Attention; Renewable Energy; Machine Learning; Time Series Forecasting

\section{Introduction}
Photovoltaic (PV) energy is increasingly gaining attention as a green and clean renewable alternative to fossil fuels. With ever increasing demands for energy from the global community, a transition from the usage of conventional fossil fuels to renewable and sustainable energy sources is necessary to address pollution and global warming, in addition to optimizing harvesting costs. However, the power generated from a PV-generation system can vary intermittently. This is attributed to a number of uncontrollable variables, such as time of day, solar panel size, local weather conditions (e.g. solar irradiation patterns, temperature, humidity, wind speed, cloud cover), accumulation of dust, and other technological limitations of the panel~\cite{SuChang,SolEngReview}. Studies show that our planet receives about $1.8 \times 10^{11} MW$ power from solar radiation at any given time, with the sun providing an energy intensity of 1376 $W/m^2$ over the atmosphere of Earth~\cite{SunData1,SunData2}. 

Therefore, a reliable and accurate method for predicting the power generated by terrestrial PV systems becomes paramount. Segmentation of cloud and satellite algorithms, numerical weather-prediction paradigms are often complicated, time-consuming, and prohibitively expensive to be relied upon for PV output prediction. More so, these methodologies are often rendered  infeasible for rural PV stations~\cite{ruralStation}. As the influence of a large number of variables can often lead to sudden, unpredicted dips or surges in power, systematic forecasting of such behaviors can help optimize the output through significant reductions in associated uncertainties of the whole subsystem, while reducing costs. Several time-series forecasting methodologies have been tried over the years to predict PV output power, but such (forecasting) pipelines often have low accuracy and throughput due to spatiotemporal fluctuations in solar irradiance~\cite{Regression2}. The majority of traditional techniques for power prediction rely on multivariate regression~\cite{Sanewal,Regression1,Regression2}, which necessitates gathering a significant number of pertinent data points, which include temperature, electricity-generation rates, and local insolation. For the purpose of time series classification, when label information for certain data points is absent, inclusion of reference data is often necessary~\cite{TimeSeries1}. In this paper, to address all these challenges, we propose SolarTformer, which can accurately capture local semantic information and benefit from the cyclic encoding we perform as time has cyclical properties. To summarize, the primary contributions of this manuscript are as follows:
\begin{itemize}
    \item[--] We propose a modified transformer~\cite{attentionall,yuqietal} based architecture for time series prediction to forecast PV output.
    
    \item[--] Since the temporal attribute exhibits cyclic patterns, the time series data obtained from solar power stations across China are cyclically encoded to ensure efficient model convergence
    
    \item[--] We integrate metadata from the time series into the deep neural architecture, addressing limitations in existing approaches by capturing intrinsic dataset characteristics while significantly enhancing the prediction performance over pre-existing methods. This improves generalization across unseen weather conditions as reported in the improved accuracy and lower mean prediction error. Our results demonstrate that SolarTformer surpasses existing literature by a significant margin and achieves a decrease in mean percentage error by almost 60\%.
\end{itemize}

The remaining manuscript is structured as follows. A brief overview of recent, relevant published works is presented in Section~\ref{related-works}. This is followed by a description of the dataset, data preprocessing, and the proposed deep learning model in Section~\ref{methodology}. Then, Section~\ref{results} discusses the results and provides a comparison with other works on the same dataset, and relevant ablation studies. Finally, the manuscript ends with 'Conclusions and Future Scope' in Section~\ref{conclusion}, where we discuss future work and how to improve on certain limitations. We provide pseudocode for all algorithms used in our work in the Appendix for reproducibility.

\section{Related Works}
\label{related-works}
Earlier published works show that research on PV-generated power prediction is sparse, as most literature focus on irradiance patterns forecasting~\cite{Sharadga2019}.
In published research literature dealing with prediction of output power , temporal data forecasting using statistical methods have been largely used, as they do not require information on the physical aspect of power generation in PV systems, but leverage previous data collected from plant sites directly~\cite{Yurui2023}. The most commonly employed statistical linear approaches are moving average filter models, auto-regressive models, and their derivatives such as the auto-regressive moving-average (ARMA) time series models which has high accuracy on short-term prediction jobs~\cite{Feng2020,Erdem2011}. Linear models, even though simple to implement, has drawbacks for being non-flexible with high-dimensional data, such as obtained from photovoltaic plants, which vary according to sporadic weather conditions. Yanting et al.~\cite{Yanting2016} have used a non-linear approach with multivariate adaptive regression splines (MARS) to predict daily output power, where the model does not explicitly need any assumption about dependencies between the power output and other variables. 

These statistical methods, even though largely explainable and interpretable, fail to predict with accuracy, and introduce large error rates when encountered with intermittent changes in PV power generation, as it is susceptible to weather change and variations in solar positions over the day~\cite{Xiang2024}.

Over the last few decades, several machine learning models have been implemented where, more often than not, manual feature extraction was necessary. Abstracting away the simplicity of linear models, artificial neural networks presents a trade-off between complex paradigms and larger computational costs. With the advent of large scale, economically viable GPU clusters, this has not been the case recently, where deep learning architectures such as the Long Short Term Memory (LSTM)~\cite{Hossain2020,Tang2021}, Gated Recurrent Units (GRU), Recurrent Neural Networks (RNN)~\cite{Stan2021}, and the Elman neural networks have been heavily relied on for PV output prediction tasks.
Shakhovska et al.~\cite{Shakhovska} have applied machine learning (ML) and deep learning (DL) techniques to predict solar energy generation, highlighting the importance of detailed meteorological data. They have collected and preprocessed comprehensive datasets, introduced an hour-based prediction wrapper using sunrise and sunset data to focus on daylight hours, and have proposed a cascaded stacking model to improve generalization and errors minimization. Models trained on raw data have outperformed those on stripped data, with an LSTM-Inception model achieving the best results. The study shows that combining rich weather data with advanced ML/DL methods can significantly enhance solar energy forecasting.
Chaaban et al.~\cite{Chaaban} have compared various ML models, including the Artificial Neural Network (ANN), the Random Forest, and the LSTM, for predicting PV energy yield using weather forecast data. They found that the Random Forest consistently achieved the best performance, with Mean Absolute Error (MAE) and Root Mean Squared Error (RMSE) improving further when combined with solar irradiation data.
Ferreira et al.~\cite{Ferreira} have applied ML to predict optical properties of materials for building-integrated photovoltaics (BIPV) using photo-luminescent measurements. Regression and clustering models estimated optical and power conversion efficiencies with the MAEs as low as 7\%, offering a cost-effective, efficient alternative to traditional material design methods.
Ledmaoui et al.~\cite{Ledmaoui} compared six ML algorithms (ANN, Support Vector Regression, Decision Tree, Random Forest, Generalized Additive Model and the Extreme Gradient Boosting) for solar energy forecasting in Morocco. Their findings have revealed ANN to outperform others, with the lowest RMSE and highest R-squared. The study identifies optimal ANN configurations, offering practical insights for improving real-world solar energy system optimization.
Ganthia et al.~\cite{Ganthia} proposed a smart energy harvesting model (MEHMM), analyzing how reflection varies with the angle of incidence on glass surfaces. Compared to existing models such as the optimized fuzzy logic control (OFLC), fuzzy logic control of stand-alone energy harvesting solar photovoltaic system (FLCSP), an improved fuzzy logic controller design (IFLCD), and the fuzzy probabilistic-based semi-Markov model (FPSMM), the MEHMM more efficiently sensed and transmitted side light, improving energy capture on non-smooth surfaces.
Park et al.~\cite{Park} used ML models to predict the output of hybrid energy devices (HEDs) that combines photovoltaic cells and thermoelectric generators. Among the eight models, the ANN best captured the relationship between interface materials and performance, showing a 2.6\% power increase (as compared to that of a photovoltaic cells alone) with a carbon paste interface at 1000 W/m².
Ramadhan et al.~\cite{Ramadhan} have compared physical and ML models for each step of solar power modeling. The ML models generally outperformed physical models, significantly reducing mean bias difference (MBD) for global irradiance and PV power, with the LSTM and the GRU performing best. However, the physical models were more effective in minimizing root mean square difference.
Vatti et al.~\cite{Vatti} developed a solar tracking system using micro-controllers and ML to optimize panel orientation year-round, following the annual motion of the sun. The proposed system have achieved 93\% training and 94\% validation accuracy. The study also highlights the importance of phase-changing materials (PCMs) for efficient solar thermal energy storage and minimizing energy loss due to reflection.
In recent literature multiple hybrid models ~\cite{CNNGRU,LSTMCNN,CNNLSTM} have also been implemented to address several limitations such as overfitting, gradient vanishing and ever-increasing complexities between the variables associated with historical data available from power plants. 

Most of the ML and DL based work, in present literature, have focus on PV prediction on similar days (i.e. days having similar weather conditions). This hinders generalizability as it requires, first sorting days based on weather conditions; and then training different model. Our proposed work aims to fill in this gap by creating a generalized model to work across different weather conditions year-round. Additionally, imporving geographical and station based generalizabilty have also been attempted by incorporating station based metadata.

\section{Methodology}
\label{methodology}
\subsection{Data Preparation}
In this manuscript, meteorological data collected from an actual PV station, along with other necessary information on solar irradiance, wind speed, environmental conditions, and mechanical characteristics (such as panel size, load, etc.) were used. The PV power output dataset (PVOD) thus used is taken from a published study~\cite{PVOD}, and contains PV power output based on numerical weather prediction (NWP) as well as local measurements data (LMD) for ten PV systems stationed in the Hubei Province of China. The data was collected with 15 minutes resolution. The dataset also contains a metadata file, which describes the characteristics of the PV power stations (see Table~\ref{tab:meta}), particularly useful for our approach.

\begin{table}[h!]
    \setlength{\tabcolsep}{4pt}
    \renewcommand{\arraystretch}{1} 
    \caption{Metadata description}
    \centering
    \begin{tabular}{l|r}
         \hline
         Data & Unit \\
         \hline
         ID of stations & \\
        capacity of the power station & kW \\
         Material type of PV panel & \\
        Size of a PV panel  & $m^2$ \\
        Module information of the PV panel & \\
        Solar inverters information of the PV system & \\
        Total number of PV panels laid in for the station & \\
        Angle of PV panels & degree \\
        Pyranometers information of the station & \\
        Longitude of the station & degree \\
        Latitude of the station & degree \\
         \hline
    \end{tabular}
    \label{tab:meta}
\end{table}

\begin{table}[h!]
    \setlength{\tabcolsep}{4pt} 
    \caption{Station Dataset Description}
    \centering
    \begin{tabular}{p{4.2cm}|r}
         \hline
         Data & Unit \\
         \hline

        date\_time format & YYYY-MM-DD Hour-Min \\
        global irradiance of NWP & $Wm^{-2}$ \\
        direct irradiance of NWP & $Wm^{-2}$ \\
        10-meter dry-bulb temperature of NWP & $^{\circ}C$ \\
        10-meter relative humidity of NWP & \% \\
        10-meter wind speed of NWP & $ms^{-1}$ \\
        10-meter wind direction of NWP, zero north clockwise & degree \\
        atmospheric pressure of NWP & hPa \\
        global irradiance of LMD & $Wm^{-2}$ \\
        diffuse irradiance of LMD & $Wm^{-2}$ \\
        temperature of LMD & $^{\circ}C$ \\
        atmospheric pressure of LMD & hPa \\
        wind direction of LMD & degree \\
        wind speed of LMD & $ms^{-1}$ \\
        PV output of the station & MW \\
         \hline
    \end{tabular}
    \label{tab:stn}
\end{table}

The power values were used as the output, and the LMD values from the station dataset (see Table~\ref{tab:stn}) were used as input values. This input dataset is referred to as "weather dataset" in the subsequent portions of this manuscript. From the metadata file, the size and number of panels, the angle of the panels, and the longitude and latitude of the power station were used as inputs. This dataset is referred to as the "meta dataset" in the remaining part of this manuscript. The aim of the present work is to use input data at time instant $t\,\text{min}$, along with the necessary metadata, in order to forecast the power output at $t + 15\,\text{min}$.

Cyclic encoding was used to represent the day and time fields. The rationale behind selecting the cyclic encoding tool is that weather and climatic conditions reveal the cyclic nature, based on time and day. For example, the ambient temperature (in general) is similar at 23:00, 0:00 and 1:00 hours. Similarly, the climatic conditions are usually similar on the day 365 of a year, and the day 1 and day 2 of the next year. Using linear measurements (like 23:00, 0:00, 1:00 for time and 365, 1, 2 for day) puts disproportionately large differences between similar values.
For cyclic encoding, the data points are represented (day or time) as points on the circumference of a circle and, thus, can be represented by the parametric equation of a circle $(cos(\theta), sin(\theta))$ (assuming radius = 1).
If a data (of cyclic nature) has, say, $x$ number of data points, each point can be thought to be on the circumference separated by an angle of $\frac{2\pi p_i}{x}$, where $p_i$ is the $i^{th}$ data point (assuming the data points are evenly spaced).
\\
Thus, the cyclic encoding for the day and time values can be represented as:
\begin{equation} \label{eq:time-cyclic}
time_{i\_cyclic} = \cos{(\frac{2 \pi  time_i}{96})}, \sin{(\frac{2 \pi  time_i}{96})}
\end{equation}

\begin{equation} \label{eq:day-cyclic}
day_{i\_cyclic} = \cos{(\frac{2 \pi  day_i}{365})}, \sin{(\frac{2 \pi  day_i}{365})}
\end{equation}

In Eq.~\ref{eq:time-cyclic}, $time_i$ refers to the $i^{th}$ time interval and $time_{i\_cyclic}$ refers to the corresponding cyclic encoding. Data was collected at 15 min interval, which makes a total of 96 times a day when data was collected. This appears as the denominator in Eq. \ref{eq:time-cyclic}.
Similarly, in Eq.~\ref{eq:day-cyclic}, $day_i$ refers to the $i^{th}$ day of the year (with $1^{st}$ January being day 1 and $31^{st}$ December being day 365, for non-leap years) and $day_{i\_cyclic}$  refers to the corresponding cyclic encoding.
\begin{figure} 
    \centering
    \includegraphics[width=0.45\textwidth, height = 7cm]{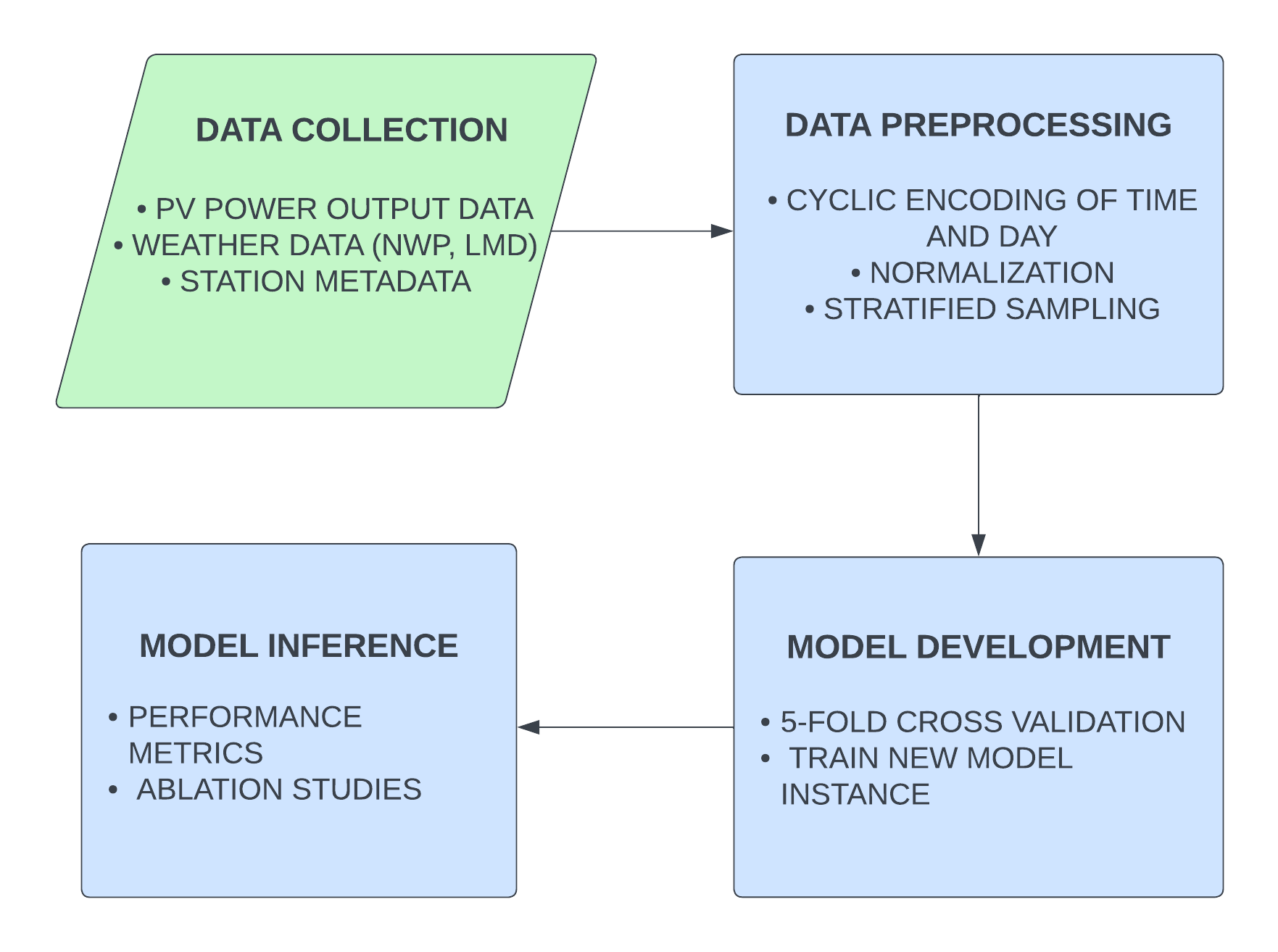}
    \caption{Training Workflow}
    \label{fig:training-workflow}
\end{figure}
The data set contained data measured over the time span 2018-2019 that does not contain any leap year, and thus, it is sufficient to only use 365 as the denominator in Eq.~\ref{eq:day-cyclic}. The initial weather dataset was of shape [day-time, weather values]. In this study, it was reshaped into [day, time, weather value]. Similarly, the power output dataset was of shape [day-time, power value], and it was reshaped to [day, time, power value]. The power values were normalized by dividing by the number of solar panels in the respective stations. 
The original meta data set contained a single row corresponding to each PV station.  In this work, it was increased to match the number of rows in the corresponding weather dataset, i.e. one row for each day. Along with this, the cyclic encoding for day was added in the meta dataset. This was done to feed the meta data along with the weather data to our neural network. The datasets of the individual stations were joined (row wise) to create the final dataset. The final dataset, thus, contained 2515 records. The dataset was split into training and testing datasets in a ratio of 9:1. 

The training dataset was used to do 5-fold cross validation. All the dataset splits were done using Stratified Random Sampling, with the station as the strata, to ensure that similar statistical parameters are maintained across the different dataset splits. These datasets were normalized using the z-score normalization, as this type of scaling is more robust to outliers. During scaling, it was also ensured that the cyclic encoded values remained unaffected.

\subsection{Model Architecture}
Our proposed power forecasting model, the "SolarTformer", utilizes a hybrid transformer architecture designed for time-series prediction, integrating multivariate weather data and static metadata. The input consisted of a weather sequence of 96 time steps, embedded via a ReLU-activated dense layer into a 64-dimensional space, and static metadata projected into the same dimensionality.
\begin{figure*}
  \centering
  \includegraphics[width=0.8\textwidth, height=11cm]{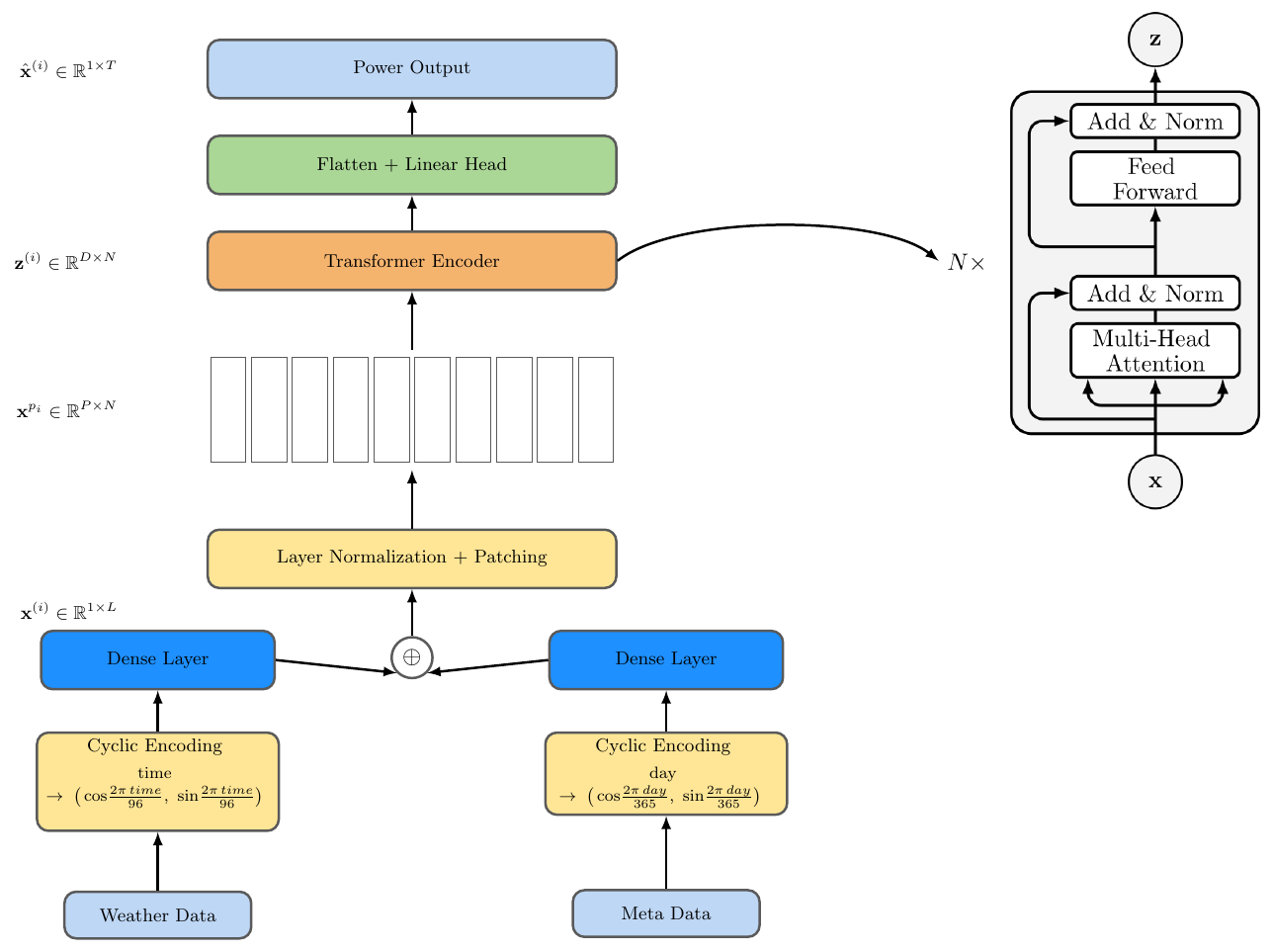}
  \caption{SolarTformer Architecture}
  \label{fig:network-architecture}
\end{figure*}
A trainable start token S was concatenated with the weather embeddings along the temporal axis. The metadata embedding was tiled across all time steps and concatenated with the weather embeddings, forming the input W. The model enforces temporal dependencies using a causal mask to enforce a fundamental constraint: predictions at any time point depend only on the information available up to that moment, ensuring temporal causality. Mathematically, the causal mask is defined as follows:

\[
M_{\text{causal}}[i, j] = \begin{cases} 
0, & j \leq i \\ 
-\infty, & j > i 
\end{cases}
\]

This matrix structure ensures that each element \((i,j)\) is zero when position \( j \) is at or before position \( i \), allowing attention, while it is \( -\infty \) if position \( j \) is in the future, thereby blocking attention. When applied in the attention mechanism, large negative values translate to near-zero attention weights after softmax, eliminating any contribution from future time steps.

In the context of power forecasting, this causal constraint ensures that the power prediction at timestep \( t \), denoted as \( p_t \), is computed solely based on past weather data and static metadata, expressed as:

\[
p_{t} = f(w_{< t}, m)
\]

where \( f \) represents the learned mapping of the transformer, \( w_{< t} \) denotes the weather characteristics up to time \( t \) (in this case, up to 15 min before time \( t \)), and \( m \) represents the static metadata. Without this causal masking mechanism, the model could access future weather data, leading to information leakage and an unrealistic forecast scenario. The mask is integrated within the self-attention mechanism, ensuring strict adherence to causality. This can be expressed as:

\begin{multline}
\text{Attention}(Q, K, V, M_{\text{causal}}) \\
= \text{softmax}\left(\frac{QK^T + M_{\text{causal}}}{\sqrt{d_k}}\right)V
\end{multline}

where \( Q \), \( K \), and \( V \) are the query, key, and value matrices derived from weather and metadata embeddings, and \( d_k \) represents the dimension of the key vectors. Our architecture employs multi-head self-attention layers, each with four attention heads operating in 16-dimensional subspaces.

The model was trained using the mean squared error (MSE) loss and AdamW~\cite{loshchilov2018decoupled} was chosen as the optimizer with an initial learning rate of 0.01, with performance evaluated using Concordance Correlation Coefficient (CCC), Percentage Error (PE), and Kullback-Leibler Divergence (KLD). 5-fold cross validation was done with 5 different instances of the model to assess the applicability and capacity of the proposed model to learn the probability distribution. All the model instances were trained for 200 epochs each during cross validation. Afterward, a new instance of the model was trained on the entire training set for 300 epochs and evaluated on the test set. The whole training pipeline is depicted in Fig.~\ref{fig:training-workflow}.

\section{Results}
\label{results}
The results of the proposed model are presented in Tables \ref{tab:cv} and \ref{tab:results}. Table \ref{tab:cv} reports the mean squared error (MSE) values across different cross-validation folds. The relatively consistent values across the folds demonstrate the model’s stability and robustness during training. This consistency indicates that the model generalizes well, and does not heavily rely on specific data subsets.

However, when trained without regularization, signs of slight overfitting was exhibited by the model, as indicated by the larger discrepancy between the training and the validation errors. To mitigate this, elastic net regularization was applied with both L1 and L2 coefficients set to $10^{-4}$. This modification has resulted in a reduced gap between the training and the validation errors, suggesting improved generalization performance.

\begin{table*}[ht]
\setlength{\tabcolsep}{10pt}
\renewcommand{\arraystretch}{1.5}
\caption{Mean squared error across cross-validation folds, with and without regularization}
\centering
\begin{tabular}{|l|c c|c c|}
\hline
\textbf{Fold} & \multicolumn{2}{c|}{\textbf{Without Regularization}} & \multicolumn{2}{c|}{\textbf{With Regularization}} \\
\cline{2-5}
 & Training & Validation & Training & Validation \\
\hline
1 & 1.0294 & 1.8963 & 1.7265 & 1.7962 \\
2 & 1.1304 & 1.3988 & 1.5472 & 1.5824 \\
3 & 1.1538 & 1.4889 & 1.6177 & 1.6899 \\
4 & 1.1249 & 2.1716 & 1.1012 & 1.2560 \\
5 & 1.2706 & 1.5549 & 1.5391 & 1.5814 \\
\hline
\textbf{Mean} & 1.1418 & 1.7021 & 1.5063 & 1.5812 \\
\hline
\end{tabular}
\label{tab:cv}
\end{table*}

\subsection{Evaluation Metrics}

In addition to the MSE, other metrics were used to evaluate the model's performance on the test dataset, including the Percentage Error (PE), the Kullback–Leibler (KL) Divergence, and the Lin's Concordance Correlation Coefficient (CCC).

\textbf{Mean Absolute Percentage Error (MAPE):}
The MAPE is defined as:
\begin{equation}
\text{MAPE} = \frac{1}{n} \sum_{i=1}^{n} \left| \frac{y_i - \hat{y}_i}{y_i} \right| \times 100
\end{equation}
where $y_i$ is the true value, $\hat{y}_i$ is the predicted value, and $n$ is the number of data points.

Using the above formula of MAPE was infeasible, as it contained several $y_i$ that had a 0 value, resulting in 'division by zero' error. Thus, an alternate way of calculating the Percentage Error (PE) has been used, described as follows:
\begin{equation}
    \text{PE} = \frac{1}{n} \frac{\sum_{i=1}^{n} \left| y_i - \hat{y}_i \right|}{\sum_{i}^{n} \left| y_i \right|} \times 100
\end{equation}
This is in line with the error metric used in the published literature~\cite{amartySolar,LiuErr} on the same dataset. Unlike traditional MAPE, this formulation remains well-behaved even when $y_i \to 0$ making it better suited to low-irradiance or zero power output conditions.

\textbf{KL Divergence:}
The KL divergence is defined as:
\begin{equation}
D_{\text{KL}}(P \| Q) = \sum_{i} P(i) \log \frac{P(i)}{Q(i)}
\end{equation}
where $P(i)$ and $Q(i)$ are the true and predicted probability distributions, respectively.

\textbf{Lin's Concordance Correlation Coefficient (CCC):}
The CCC is defined as:
\begin{equation}
\rho_c = \frac{2 \rho \sigma_x \sigma_y}{\sigma_x^2 + \sigma_y^2 + (\mu_x - \mu_y)^2}
\end{equation}
where $\rho$ is the Pearson correlation coefficient between the true and predicted values, $\sigma_x^2$ and $\sigma_y^2$ are the variances, and $\mu_x$ and $\mu_y$ are the means of the true and predicted values, respectively.

From Table \ref{tab:results}, it can be seen that the proposed model shows very good predictive performance with a low 1.7312\% error, 0.0232 KL divergence and high CCC value of 0.9764.

\begin{table}[!h]
\setlength{\tabcolsep}{3pt}
\renewcommand{\arraystretch}{0.9}
\caption{Test dataset evaluation metrics}
\centering
\resizebox{0.7\linewidth}{!}{
\begin{tabular}{|c|c|c|c|}
\hline
\textbf{MSE} & \textbf{PE} & \textbf{KL Divergence} & \textbf{CCC} \\
\hline
1.6438 & 1.7312 & 0.0232 & 0.9764 \\
\hline
\end{tabular}
}
\label{tab:results}
\end{table}

\begin{figure}[ht]
\centering
\includegraphics[width=0.47\textwidth, height = 4.9cm]{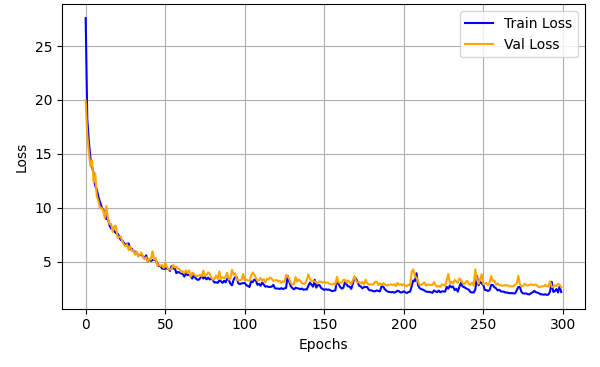}
\caption{Loss curve for the final model training over 300 epochs}
\label{fig:training_loss}
\end{figure}

The final model was trained for 300 epochs. Till then, the error had plateaued (i.e., there was no appreciable reduction in the loss in sucssive epochs). This can be seen in Figure~\ref{fig:training_loss} which shows the variation of the loss function (for both training and validation on the test dataset) against the epoch number.

\vspace{-3pt}
\subsection{Ablation Study}
An ablation study was also conducted to systematically investigate the contribution of various architectural components to the overall performance of the proposed model. The analysis focuses on how each modification affects the model's ability to generalize, as measured by the MSE across the training, validation, and test datasets.

The results are presented in Table~\ref{tab:ablation}. As discussed before, the datatset was split into training and testing sets in the ratio 9: 1. For this ablation study, the previous training data set was further split into training and validation sets in the ratio 8:2.

\begin{table}[!ht]
\small
\setlength{\tabcolsep}{4pt}
\renewcommand{\arraystretch}{1}
\caption{Ablation Study Metrics}
\centering
\resizebox{\linewidth}{!}{
\begin{tabular}{|l|c|c|c|}
\hline
\textbf{Setting} & \textbf{Train MSE} & \textbf{Validation MSE} & \textbf{Test MSE} \\
\hline
Without Normalization & 4.772 & 4.368 & 4.281 \\
Without Metadata & 2.326 & 2.431 & 2.395 \\
Without Skip Connections & 2.938 & 2.754 & 2.646 \\
Without Attention layers & 34.746 & 35.042 & 30.921 \\
$N = 1$ & 2.212 & 2.694 & 2.201 \\
$N = 2$ & 1.145 & 1.561 & 1.644 \\
$N = 4$ & 1.604 & 2.033 & 1.802 \\
$N = 6$ & 1.825 & 2.531 & 1.729 \\
\hline
\end{tabular}}
\label{tab:ablation}
\end{table}

The findings clearly demonstrate the critical role of several key components in achieving high predictive accuracy. Removing normalization layers results in a significant degradation in performance, as compared to other settings. This highlights the importance of normalization in stabilizing training, and ensuring consistent model behavior across different data distributions. Similarly, omitting metadata features also leads to an increase in both validation and test errors. This indicates the importance of including metadata in the input for better predictive performance.

The removal of skip connections, likewise, adversely impacts performance, indicating that these connections facilitate better gradient flow and information propagation across the network. Models without skip connections exhibit higher errors, confirming their necessity for maintaining a robust learning process.

The impact of attention layers is particularly pronounced. Without attention mechanisms, the model suffers a dramatic increase in error across all datasets, underscoring the crucial role played by the attention layers in enabling the model to focus on the most relevant features and temporal dependencies.

Finally, varying the number of network blocks (where each block consists of a transformer encoder, as shown in Fig. \ref{fig:network-architecture}), denoted by parameter $N$, reveals certain interesting trade-offs. While using $N = 1$ or $N = 2$ provides relatively low training errors, the configuration with $N = 2$ achieves the best balance between the validation and the test performance. Increasing $N$ beyond 2, particularly to $N = 4$ or $N = 6$, leads to diminishing returns or even slight performance degradation on the test set, which is likely due to the increased model complexity and the potential for overfitting.

In summary, the ablation study validates the design choices of the proposed model architecture. Each component contributes meaningfully to the model's predictive performance, and careful tuning of the parameters is essential for optimizing performance. The insights gained from this analysis offer avenues for future improvements and demonstrate the robustness of the model's architecture.

\subsection{Comparison with Other Models}

The performance of our proposed model is compared with other baseline methods, in published literature, used on the same dataset, as shown in Table~\ref{tab:res-comparison}. Unlike prior works, which trained separate models for sunny and non-sunny days, our unified model is trained on all types of weather conditions and still performs competitively. Also, in the present work, only one model was trained for data across all the seasons and our model was found to perform well. This can be attributed to the inclusion of seasonal information into the meta-data via the cyclic-encoding. 

\begin{table}[!ht]
\setlength{\tabcolsep}{2pt}
\renewcommand{\arraystretch}{1}
\caption{Comparison of percentage error across models}
\centering
\resizebox{\linewidth}{!}{
\begin{tabular}{p{8cm}r}
\hline
\textbf{Model} & \textbf{Percentage Error} \\
\hline
SVM \cite{helmySolar} & 45.00 \\
Generalized Spatio-Temporal Artificial Neural Network (GSTANN) ~\cite{helmySolar,yaoSolar} & 38.00 \\
CNN (Average sunny + non sunny days)\cite{pengSolar} &  16.57\\
BP (Average sunny + non sunny days) \cite{pengSolar} & 11.62 \\
LSTM (Average sunny + non sunny days)\cite{pengSolar} & 6.65\\
Similar Day 1DCNN-LSTM + LightGBM (SD-CL-LGBM) (Average sunny + non sunny days) ~\cite{pengSolar} & 5.17\\
GRU \cite{YuanSolar} & 2.44 \\
LSTM-CNN \cite{YuanSolar} & 2.62 \\
GRU + multi-timescale fluctuation aggregation (MFA) Attention \cite{YuanSolar} & 2.41 \\
LSTM-CNN + MFA Attention \cite{YuanSolar} & 2.63 \\
Similar Day - Backpropagation Neural Network + AdaBoost - Grey-Markov (SD-BPA-GM) (Non Sunny) \cite{YangSolar} & 4.84 \\
SD-BPA-GM (Non Sunny) \cite{YangSolar} & 8.49 \\
Hybrid Inception-ResNet + Informer (Spring) \cite{pengSolarInception} & 8.81\\
Hybrid Inception-ResNet + Informer (Summer) \cite{pengSolarInception} & 8.17\\
Hybrid Inception-ResNet + Informer (Autumn) \cite{pengSolarInception} & 6.95\\
Hybrid Inception-ResNet + Informer (Winter) \cite{pengSolarInception} & 5.80\\
TinyML \cite{amartySolar} & 4.51\\
\textbf{SolarTformer} & \textbf{1.73} \\
\hline
\end{tabular}
}
\label{tab:res-comparison}
\end{table}

\begin{figure*}[t]
\centering

\subfloat[]{\includegraphics[width=0.47\textwidth]{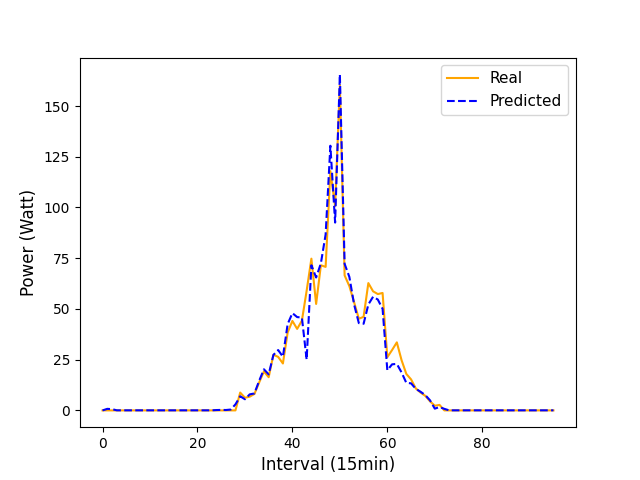}}
\hfill
\subfloat[]{\includegraphics[width=0.47\textwidth]{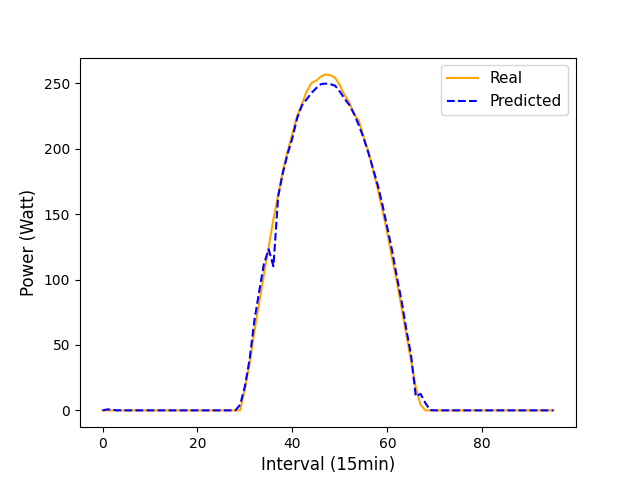}}

\subfloat[]{\includegraphics[width=0.47\textwidth]{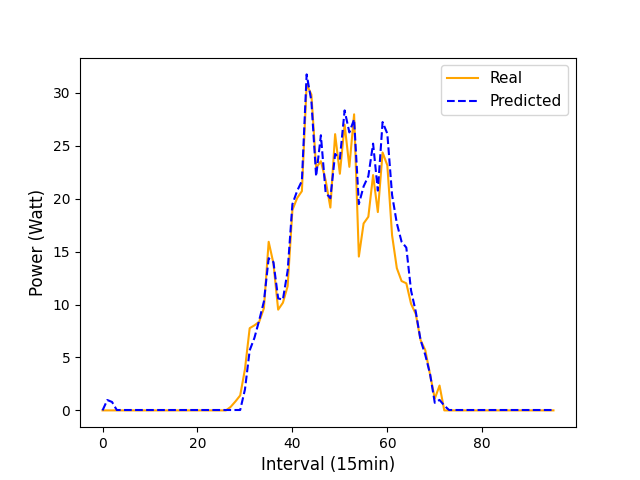}}
\hfill
\subfloat[]{\includegraphics[width=0.47\textwidth]{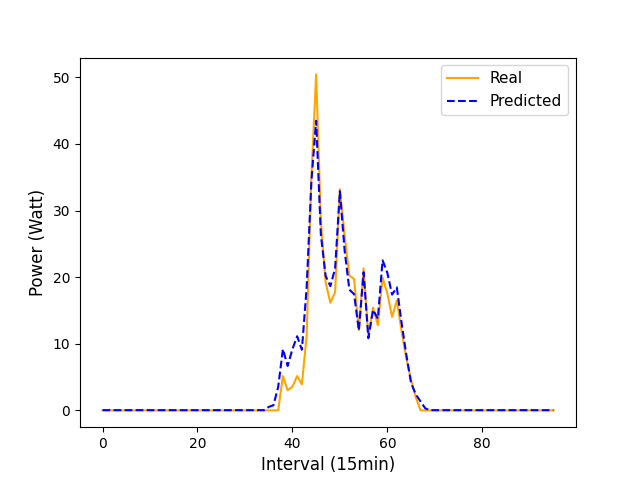}}

\subfloat[]{\includegraphics[width=0.47\textwidth]{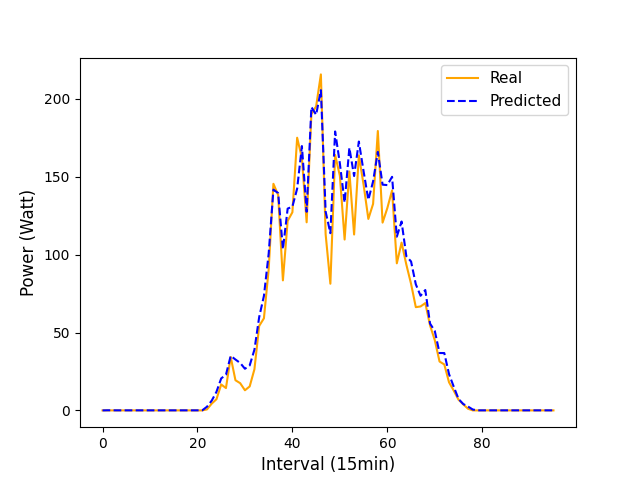}}
\hfill
\subfloat[]{\includegraphics[width=0.47\textwidth]{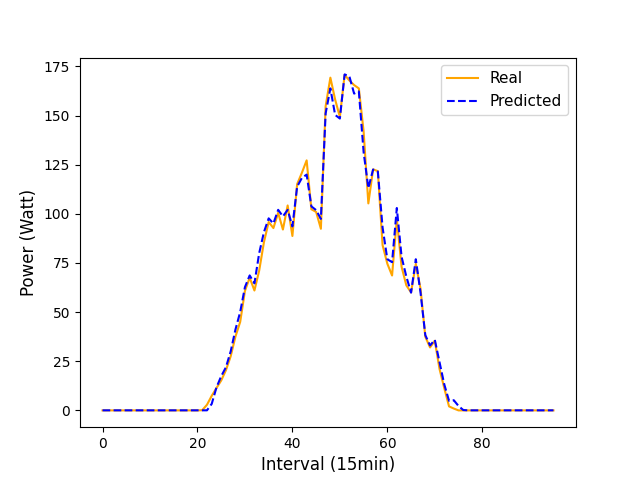}}

\caption{Power predictions across different test samples}
\label{fig:test_predictions}
\end{figure*}

The strong performance of our proposed model across various weather conditions can be seen in Fig. \ref{fig:test_predictions} that depicts the plot of generated power versus time for some data points (days) in the test dataset. From the figure, it can be seen that the predicted power output closely matches the real data across multiple days. The plots with high power output typically correspond to sunny days with strong solar irradiation, while low output indicates overcast or less sunny conditions. Moreover, a smooth power curve throughout the day suggests uninterrupted sunlight, whereas fluctuations with multiple peaks and troughs reflect intermittent cloud cover. The plots in Fig. \ref{fig:test_predictions} show the close alignment between our model’s predictions and actual outputs under both stable and varying weather conditions. This demonstrates the robustness and generalizability of SolarTformer.

\begin{figure*}[t]
\centering

\subfloat[]{\includegraphics[width=0.47\textwidth]{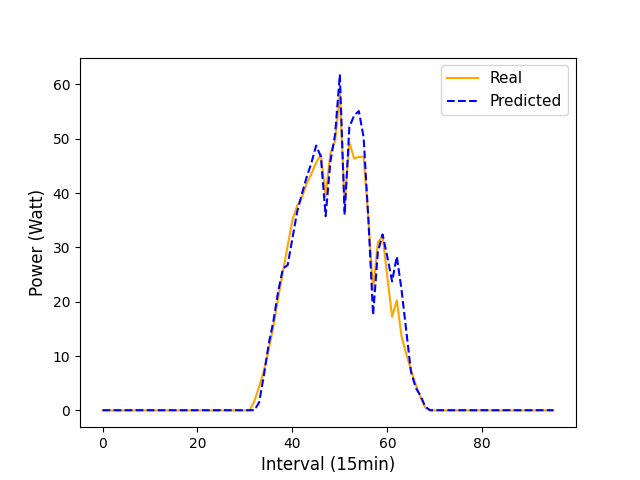}}
\hfill
\subfloat[]{\includegraphics[width=0.47\textwidth]{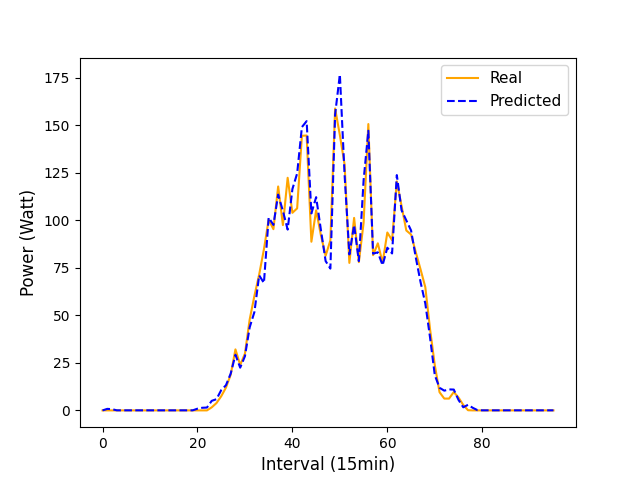}}

\subfloat[]{\includegraphics[width=0.47\textwidth]{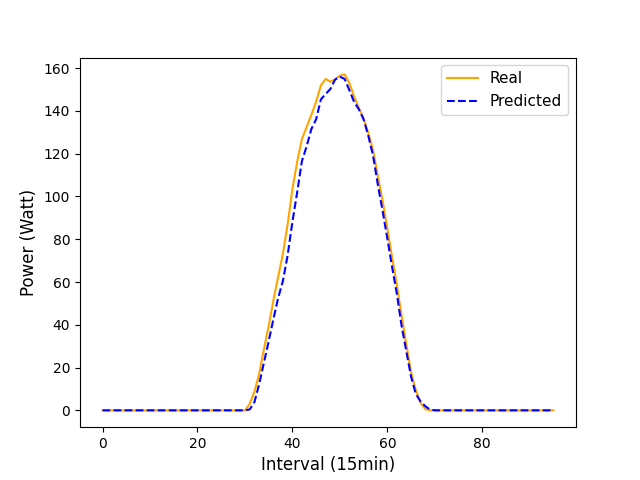}}
\hfill
\subfloat[]{\includegraphics[width=0.47\textwidth]{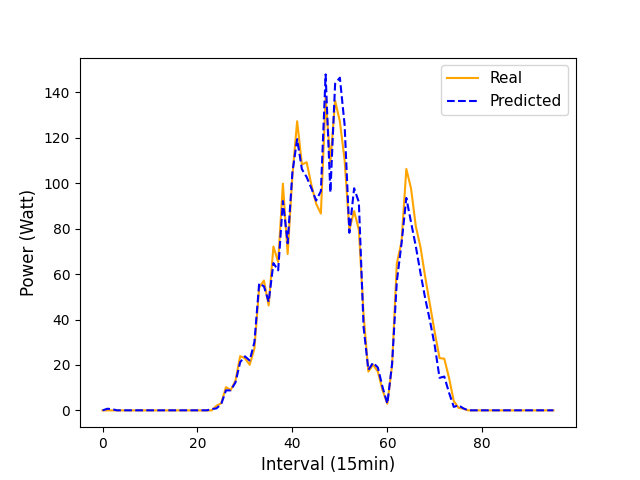}}

\caption{Power predictions across different test samples (continued).}
\end{figure*}

\clearpage

\section{Conclusion and Future Scope}
\label{conclusion}
This work presents a first step in developing an end-to-end framework for time-series forecasting on weather-based datasets, with an aim for reliable prediction of solar PV power. Owing to constraints imposed by limited computational resources, the primary focus of this work lies in demonstrating the theoretical foundation and capability of the proposed "SolarTformer" methodology to address the problem of solar power output prediction. A large-scale study on similar and more diverse datasets is planned as future scope in order to validate and scale our approach further. While "SolarTformer" has demonstrated strong predictive performance and robustness for short-term photovoltaic power forecasting, several practical and methodological limitations still remain to be addressed. First,  hyperparameters pertaining to the model were selected based on informed choices, rather than an exhaustive search, as computational constraints have prevented extensive tuning. Although the ablation study included trying out different values of $N$ (number of network blocks), other important hyperparameters such as the number of attention heads, embedding dimensions, etc. were not selected by exhaustive search. Future extension of this work is aimed to involve systematic hyperparameter optimization (e.g., via Bayesian optimization) and additional modifications to refine the architecture. 

Furthermore, despite leveraging the benefits of self-attention for ensuring temporal learning, the computational and memory demands of the employed architecture pose challenges for deployment on resource-constrained (or edge) devices. Exploring the potentials of lightweight attention approximations, model pruning, quantization, or Monte Carlo dropout would be critical for creation of efficient versions, suitable for embedded systems and rural applications. In addition, attention mechanisms may struggle with extrapolation beyond the observed data, particularly for extreme weather conditions or other unforeseen natural and/or man-made events. As pointed out, the current evaluation is limited to the dataset from Hubei Province of China spanning over 2018–2019. Validating the proposed model on geographically diverse and climatically variant datasets, especially those with unique weather patterns, is essential to assess its generalization capabilities.

Another challenge lies in the interpretability of transformer-based models. While attention weights offer a certain degree of insight into feature relevance, a deeper understanding of how specific meteorological inputs drive (power output) predictions requires advanced explainability methods (e.g., SHAP, integrated gradients). Finally, the current model focuses on a 15-minute short-term forecast horizon. Extending the present model to support multi-horizon forecasts (e.g. hourly, daily) and integrating it with real-time grid management systems would greatly broaden its practical applicability. Addressing these challenges will enhance both the theoretical underpinnings and the real-world deployment potential of the proposed model as a comprehensive time-series forecasting solution for renewable energy systems.

\section{Author Contributions}
\textbf{AB}: Conceptualization; Data curation; Formal analysis; Investigation; Methodology; Validation; Writing-original draft; Writing-review \& editing. \textbf{JR}: Investigation; Methodology; Visualization; Writing-original draft. \textbf{AD}: Investigation; Methodology; Visualization; Writing-original draft. \textbf{PS}: Conceptualization; Formal analysis; Visualization; Validation; Writing-original draft; Writing-review \& editing. \textbf{SB}: Project administration; Supervision; Validation; Writing-review \& editing.

\bibliography{refs}

\clearpage
\onecolumngrid

\appendix

\section{Supplementary Information}
\subsection{Data Preparation and Splitting}

\begin{algorithm}[H]
\caption{Data Preparation for SolarTformer}
\label{alg:app-data}

\KwIn{Station set $\mathcal{S}$; for each $s \in \mathcal{S}$: weather table $\mathsf{W}_s$ (LMD at 15-min), power table $\mathsf{P}_s$, metadata row $\mathsf{M}_s$}
\KwOut{Dataset $\mathcal{D}=\{(X_i \in \mathbb{R}^{T \times D_w},\, m_i \in \mathbb{R}^{D_m},\, y_i \in \mathbb{R}^{T},\, \text{id}_i)\}_{i=1}^{N}$, with $T{=}96$}

$\mathcal{D} \gets \emptyset$\;

\For{each station $s \in \mathcal{S}$}{
  Parse \texttt{date\_time} into day-of-year $d \in \{1,\dots,365\}$ and time-slot $\tau \in \{0,\dots,95\}$\;

  Reshape weather: $X_s[d,\tau,:] \in \mathbb{R}^{T \times D_w}$ from $\mathsf{W}_s$ (one row per day)\;
  Reshape power: $Y_s[d,\tau] \in \mathbb{R}^{T}$ from $\mathsf{P}_s$ (one row per day)\;

  Per-panel normalization: $Y_s \gets Y_s / \texttt{num\_panels}(\mathsf{M}_s)$\;

  Cyclic time features for each $\tau$: $t_{\cos}(\tau){=}\cos(2\pi\tau/96)$, $t_{\sin}(\tau){=}\sin(2\pi\tau/96)$\;
  Cyclic day features for each $d$: $d_{\cos}(d){=}\cos(2\pi(d{-}1)/365)$, $d_{\sin}(d){=}\sin(2\pi(d{-}1)/365)$\;

  Expand metadata to daily rows and append day encoding:\;
  $m_s[d]{:=}\text{concat}\!\big(d_{\cos}(d),\; d_{\sin}(d),\; \mathsf{M}_s\big) \in \mathbb{R}^{D_m}$\;

  \For{each day $d$ present in $X_s, Y_s$}{
    Augment weather time-axis with time encodings:\;
    $X^{\star}_s[d,\tau,:] \gets \text{concat}\!\big(t_{\cos}(\tau),\; t_{\sin}(\tau),\; X_s[d,\tau,:]\big)$\;

    Append sample to $\mathcal{D}$:\;
    $\mathcal{D} \gets \mathcal{D} \cup \{(X_i{:=}X^{\star}_s[d,:,:],\, m_i{:=}m_s[d],\, y_i{:=}Y_s[d,:],\, \text{id}_i{:=}s)\}$\;
  }
}

Stratified split by station ID: $\mathcal{D}^{\text{train}}, \mathcal{D}^{\text{test}}$ with ratio $9{:}1$\;

Fit z-score scaler $\phi$ on $\mathcal{D}^{\text{train}}$ weather dimensions only, excluding cyclic features $(t_{\cos}, t_{\sin}, d_{\cos}, d_{\sin})$\;

Apply $\phi$ to weather features of both $\mathcal{D}^{\text{train}}$ and $\mathcal{D}^{\text{test}}$; leave cyclic features unchanged\;

From $\mathcal{D}^{\text{train}}$, create 5 stratified folds by station ID: $\{(\mathcal{D}^{\text{tr}}_k,\mathcal{D}^{\text{val}}_k)\}_{k=1}^5$\;

\end{algorithm}

\subsection{SolarTformer Forward Pass (Causal)}

\begin{algorithm}[H]
\caption{SolarTformer Forward Pass with Causal Mask}
\label{alg:app-model}

\KwIn{Weather $X \in \mathbb{R}^{T \times D_w}$ with time encodings, metadata $m \in \mathbb{R}^{D_m}$, $T{=}96$; model dims $d{=}64$, heads $h{=}4$, blocks $N$}
\KwOut{Next-step predictions $\hat{y}_{1:T}$}

Weather embedding: $E_w \gets \mathrm{ReLU}(X W_w + b_w) \in \mathbb{R}^{T \times d}$\;

Metadata embedding: $e_m \gets \mathrm{ReLU}(m W_m + b_m) \in \mathbb{R}^{d}$\;

Trainable start token: $s \in \mathbb{R}^{d}$\;

Prepend start: $Z \gets \text{concat}([s], E_w) \in \mathbb{R}^{(T+1) \times d}$\;

Tile metadata across time: $M \gets \text{tile}(e_m, T{+}1) \in \mathbb{R}^{(T+1) \times d}$\;

Fuse modalities: $H \gets \mathrm{ReLU}\!\left([Z \, \Vert \, M] W_f + b_f\right) \in \mathbb{R}^{(T+1) \times d}$\;

Define causal mask $M_{\text{causal}} \in \mathbb{R}^{(T+1)\times(T+1)}$: $M_{\text{causal}}[i,j]{=}0$ if $j{\le}i$, else $-\infty$\;

\For{$\ell{=}1$ to $N$}{
  $U \gets \mathrm{LayerNorm}(H)$\;
  $U \gets U + \mathrm{MHA}(Q{=}U,K{=}U,V{=}U,\, M_{\text{causal}},\, h)$\;
  $V \gets \mathrm{LayerNorm}(U)$\;
  $H \gets U + \mathrm{FFN}(V)$ \tcp*{two dense layers with ReLU, residual connections}
}

Drop start token position: $H_{\text{out}} \gets H[2{:}T{+}1,:] \in \mathbb{R}^{T \times d}$\;
Readout: $\hat{y}_{1:T} \gets H_{\text{out}} W_o + b_o \in \mathbb{R}^{T}$\;

\textit{Training shift:} predict $y_t$ using $\{X_{<t}, m\}$ via the start-token shift.\;

\end{algorithm}

\subsection{Training and Cross-Validation}

\begin{algorithm}[H]
\caption{Cross-Validation Training with Elastic Net}
\label{alg:app-train}

\KwIn{Folds $\{(\mathcal{D}^{\text{tr}}_k,\mathcal{D}^{\text{val}}_k)\}_{k=1}^5$, epochs $E_{\text{cv}}{=}300$, optimizer AdamW with lr $=0.01$}
\KwIn{Loss MSE; optional elastic net: L1 $\lambda_1{=}10^{-4}$, L2 $\lambda_2{=}10^{-4}$}
\KwOut{Fold-wise train/val MSE and their means}

\For{$k{=}1$ to $5$}{
  Initialize SolarTformer parameters $\theta$\;

  \For{epoch $=1$ to $E_{\text{cv}}$}{
    \For{each minibatch $(X,m,y) \in \mathcal{D}^{\text{tr}}_k$}{
      $\hat{y} \gets \text{Forward}(X,m;\theta)$ using Alg.~\ref{alg:app-model}\;

      $\mathcal{L}_{\text{MSE}} \gets \frac{1}{|y|}\sum (y{-}\hat{y})^2$\;

      $\mathcal{L} \gets \mathcal{L}_{\text{MSE}} + \lambda_1 \lVert \theta \rVert_1 + \lambda_2 \lVert \theta \rVert_2^2$\;

      Update $\theta$ via AdamW step on $\nabla_\theta \mathcal{L}$\;
    }
    Record train MSE; compute validation MSE on $\mathcal{D}^{\text{val}}_k$\;
  }
  Store fold-$k$ train/val MSE\;
}

Report mean train/val MSE across folds, with and without regularization\;

\end{algorithm}

\subsection{Final Training, Test Time Evaluation, and Ablation Studies}
\begin{algorithm}[H]
\caption{Final Model Training and Metric Computation}
\label{alg:app-final}

\KwIn{Full training set $\mathcal{D}^{\text{train}}$, test set $\mathcal{D}^{\text{test}}$, epochs $E_{\text{final}}{=}300$, AdamW (lr $=0.01$)}
\KwOut{Test metrics: MSE, PE, KL Divergence, CCC}

Initialize new SolarTformer $\theta$\;

\For{epoch $=1$ to $E_{\text{final}}$}{
  \For{each minibatch $(X,m,y) \in \mathcal{D}^{\text{train}}$}{
    $\hat{y} \gets \text{Forward}(X,m;\theta)$\;
    Minimize MSE (and optional elastic net)\;
    Update $\theta$ via AdamW\;
  }
}

Evaluate on $\mathcal{D}^{\text{test}}$: collect $\{(y^{(i)},\hat{y}^{(i)})\}_{i=1}^{n}$ across all sequences (length $T$)\;

Compute:\;

$\mathrm{MSE} \gets \frac{1}{nT}\sum_{i,t} \big(y^{(i)}_t{-}\hat{y}^{(i)}_t\big)^2$\;

$\mathrm{PE} \gets \frac{\sum_{i,t} \lvert y^{(i)}_t{-}\hat{y}^{(i)}_t\rvert}{\sum_{i,t} \lvert y^{(i)}_t\rvert}\times 100$\;

Estimate empirical distributions $P,Q$ via histogram with $B$ bins on $\{y\}$ and $\{\hat{y}\}$; apply smoothing $\epsilon$\;

$D_{\mathrm{KL}}(P\|Q) \gets \sum_{b=1}^{B} P_b \log \frac{P_b}{Q_b+\epsilon}$\;

Let $x \gets y$ (true) and $z \gets \hat{y}$ (pred); compute means $\mu_x,\mu_z$, stds $\sigma_x,\sigma_z$, Pearson $\rho$\;

$\mathrm{CCC} \gets \frac{2\rho\,\sigma_x\sigma_z}{\sigma_x^2+\sigma_z^2+(\mu_x-\mu_z)^2}$\;

\end{algorithm}


\begin{algorithm}[H]
\caption{Ablation Protocol and Evaluation}
\label{alg:app-ablation}

\KwIn{Base pipeline (Alg.~\ref{alg:app-data}--\ref{alg:app-final}); ablations:
(i) No normalization; (ii) No metadata; (iii) No skip connections;
(iv) No attention layers; (v) $N{=}1,2,4,6$ transformer blocks}

\KwOut{Train/Validation/Test MSE for each setting}

Use the same 9{:}1 train:test split as main experiments\;

For ablation validation, further split prior training set into train:val $= 8{:}2$\;

\For{each ablation setting}{
  Modify model or preprocessing accordingly\;
  Train for $E_{\text{cv}}$ epochs on the ablation train split; validate on ablation val split\;
  Evaluate final model on test split; record Train/Val/Test MSE\;
}

Report all ablation metrics in tabular form (cf. Table~\ref{tab:ablation})\;

\end{algorithm}

\end{document}